%% file: main.tex
\pdfoutput=1 
\documentclass{article}

\usepackage{xcolor}
\usepackage[preprint]{corl_2026} 

\usepackage{graphicx}
\usepackage{booktabs}
\usepackage{wrapfig}
\usepackage{xspace}
\usepackage{pifont}
\usepackage{subcaption}
\usepackage{enumitem}
\usepackage{listings}
\usepackage{multirow}
\usepackage{array}
\usepackage{placeins}
\usepackage{tikz}
\usetikzlibrary{positioning,arrows.meta,calc,fit,backgrounds,shapes.geometric}

\input{macros}

\title{Flip Anything with Agentic Robotics Self-Learning }

\author{
   Shuangyu Xie, Kaiyuan Chen, Ken Goldberg  \\[0.2em]
   University of California, Berkeley  \\[0.2em]
\url{https://berkeleyautomation.github.io/flipanything/}
}

\begin{document}
\maketitle
\suppressfloats[t] 


\begin{abstract}
Flipping is a fundamental robotic manipulation skill that requires reorienting a target object in a stable pose on a planar support surface by rotating it 180$^\circ$ about an axis parallel to the support surface -- think of flipping a coin from heads to tails. Although flipping is a common objective in industrial tasks such as parts inspection, mechanical assembly, and tool use, coding a bimanual robot with grippers to flip diverse objects is a significant challenge.
In this paper, we show that a coding agent can autonomously develop and refine object-flipping strategies through iterative learning in physics simulations across diverse object geometries. Progressive skill acquisition, from basic handover to toppling and pushing, robust grasping and sensing, and thin-object manipulation with recovery, improves task success from 24.7\% to 95.0\%. The resulting graph-based code integrates high-level strategy selection with reactive control, enabling multi-stage flipping on physical robots.
\end{abstract}

\begin{figure*}[ht]
    \centering
    \includegraphics[width=\linewidth]{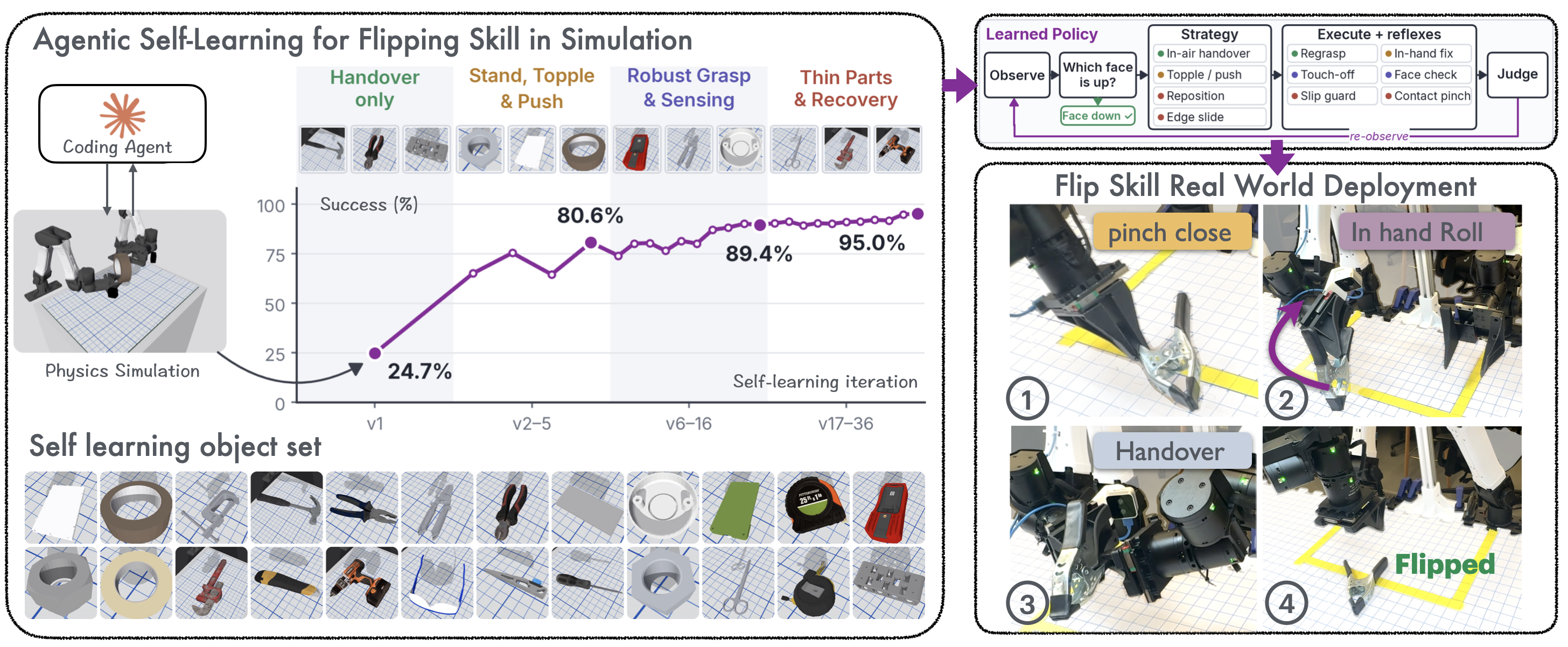}
\end{figure*}
\vspace{-1em}

\section{Introduction}
Flipping is a fundamental robotic manipulation skill that requires reorienting a target object in a stable pose on a planar support surface by rotating it 180$^\circ$ about an axis parallel to the support surface -- think of flipping a coin from heads to tails. Although flipping is a common objective in industrial tasks such as parts inspection, mechanical assembly, and tool use, coding a bimanual robot with grippers to flip diverse objects is a significant challenge.
Flipping may require diverse strategies depending on object geometry, robot kinematics, and contact constraints~\cite{chavandafle2014extrinsic,hou2018pivoting,zhou2023ungraspable,yang2024extrinsic,yamada2025combograsp}. For example, flipping a hammer may require a bimanual handover, whereas flipping a wide multimeter may require coordinated pushing and toppling. Such diversity introduces distinct action modes and coordination challenges for visuomotor policies and Vision Language Action models. 

In this short paper, we present \textit{FlipAnything}, an Agentic Robotics benchmark and case study. We use RoboSimStudio~\cite{chen2026robosimstudio} and our Graph-as-Policy~(GaP) harness~\cite{chen2026gap} with Claude Fable 5.1~\cite{anthropic2026fable}, to iteratively develop and refine an algorithmic solution through simulation-based self-learning without demonstration data. The resulting graph-based code uses RGB-D observations and robot proprioception to control bimanual YAM arms. 
We find that structured code provides several distinct multimodal manipulation behaviors, enabling a single policy to select and compose strategies such as bimanual handover, repositioning, pushing, and toppling based on perceived geometry and robot constraints. Iterative simulation-based self-learning enables the agent to identify failure modes, refine its control logic, and progressively improve coverage across diverse task instances. 
Results across 360 evaluation episodes involving 24 objects show that the graph-based code's success rate improves from 24.7\% to 95.0\%. The code learned in simulation can also transfer zero-shot to real-world manipulation of unseen objects, including a clamp, blue tape, and a foldable helix screwdriver.

\section{Method}
\textbf{Overview} 
We describe the flipping task in English and provide CAD models of 24 objects and the URDF of the YAM arms as inputs to an iterative code-generation and evaluation loop, where the coding agent improves the executable policy~\cite{liang2023codeaspolicies} through interaction with a physics simulator~\cite{fu2026capx,chen2026gap}.
We use RoboSimStudio~\cite{chen2026robosimstudio} and Newton for rigid-body physics simulation, with a MuJoCo-based solver~\cite{todorov2012mujoco} and CoACD~\cite{wei2022coacd} for convex decomposition of object meshes.
We use the GaP harness ~\cite{chen2026gap} with Claude Fable 5.1~\cite{anthropic2026fable} as the coding agent to generate execution graphs.
The agent has access to the simulator's execution context, including spatial information and execution results. 

\begin{figure}[t]
    \centering
    \includegraphics[width=0.72\linewidth]{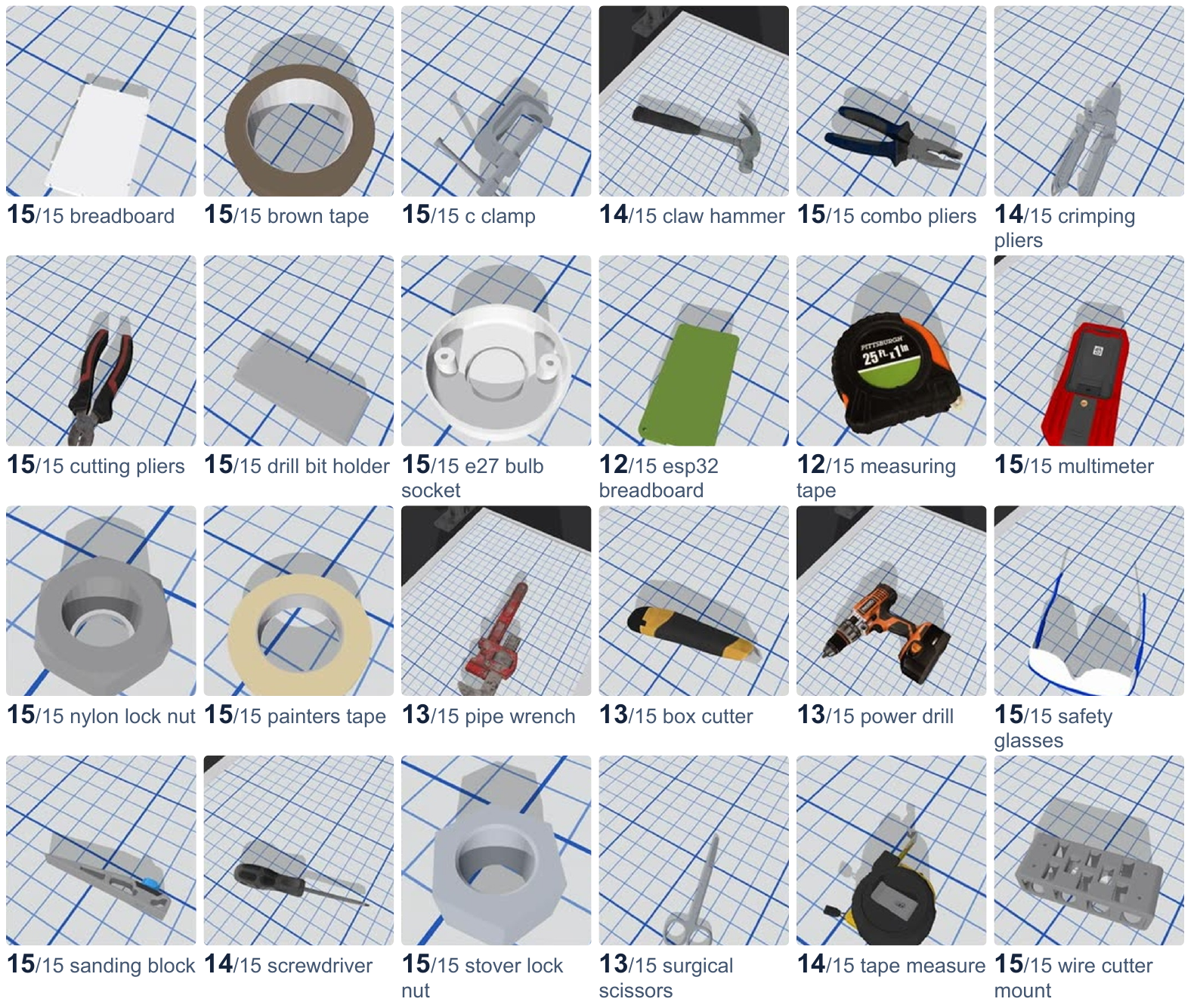}
    \caption{\textbf{FlipAnything benchmark}: 24 objects, with successes out of 15 evaluation episodes for the best tuned controller.}
    \label{fig:objects}
\end{figure}

\textbf{Agentic Self-Learning}. 
Given the flipping task and benchmark objects, the agent generates an initial graph that attempts flipping through bimanual handover. Then, it evaluates the generated flipping techniques across a set of objects with randomized positions for each object. After each evaluation, the agent analyzes success rates, execution traces, and failure modes, including infeasible grasps, lost contacts, failed handovers, and incorrect orientation estimates. Based on this feedback, the agent revises the code by refining perception, grasp selection, motion primitives, strategy selection, and recovery logic. The updated controller is then re-evaluated across the object set to measure improvements and detect regressions. Unlike gradient-based policy training, the learned artifact is a graph that can be converted into an  executable program whose conditional logic and control parameters are progressively refined to cover the 24 object geometries and manipulation strategies.

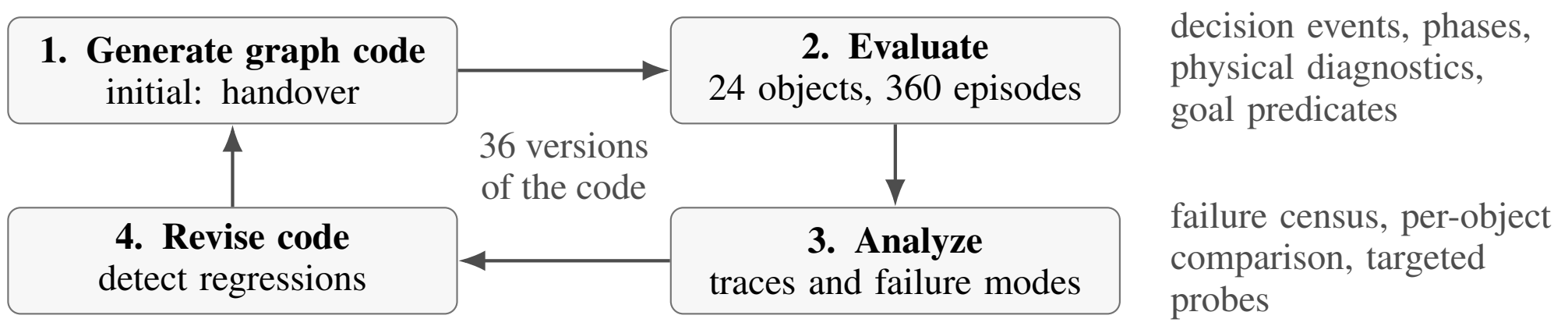
\begin{figure}[!ht]
\centering
\begin{tikzpicture}[
  font=\small, >=Latex,
  box/.style={draw=black!55, rounded corners=3pt, fill=black!3, align=center, minimum height=9mm, text width=36mm, inner sep=3pt},
  data/.style={align=left, font=\footnotesize, text=black!70, text width=33mm},
  arr/.style={->, line width=.8pt, black!70}]
\node[box] (gen) {\textbf{1. Generate graph code}\\ initial: handover};
\node[box, right=18mm of gen] (ev) {\textbf{2. Evaluate}\\ 24 objects, 360 episodes};
\node[box, below=7mm of ev] (an) {\textbf{3. Analyze}\\ traces and failure modes};
\node[box, below=7mm of gen] (rev) {\textbf{4. Revise code}\\ detect regressions};
\draw[arr] (gen) -- (ev);
\draw[arr] (ev) -- (an);
\draw[arr] (an) -- (rev);
\draw[arr] (rev) -- (gen);
\node[data, right=3mm of ev] {decision events, phases, physical diagnostics, goal predicates};
\node[data, right=3mm of an] {failure census, per-object comparison, targeted probes};
\node[font=\footnotesize, text=black!70, align=center] at ($(gen)!.5!(an)$) {36 versions\\ of the code};
\end{tikzpicture}
\caption{\textbf{Self-learning loop.} Each version is frozen before it is evaluated. The data on the right is used to revise the code and is not an input to the policy.}
\label{fig:loop}
\end{figure}

\section{Results}

\textbf{Self-Learning Improves FlipAnything Performance}.
We evaluate the agent-generated graph across 24 objects, with 15 evaluation episodes per object, totaling 360 episodes per iteration. Starting from an initial handover-only controller, the agent achieves a success rate of 24.7\% (89/360). After self-learning additional manipulation strategies, including stand-and-topple and push-based flipping, the success rate increases to 80.6\% (290/360). Subsequent self-improvement iterations refine grasp stability, geometric reasoning, contact handling, and recovery behaviors, progressively increasing performance to 95.0\% (342/360) in the best tuned controller. These results suggest that agentic robotics can improve manipulation performance by identifying failure modes and modifying its executable control logic, rather than training separate policies for different objects.

\begin{figure}[!ht]
    \centering
    \includegraphics[width=0.84\linewidth]{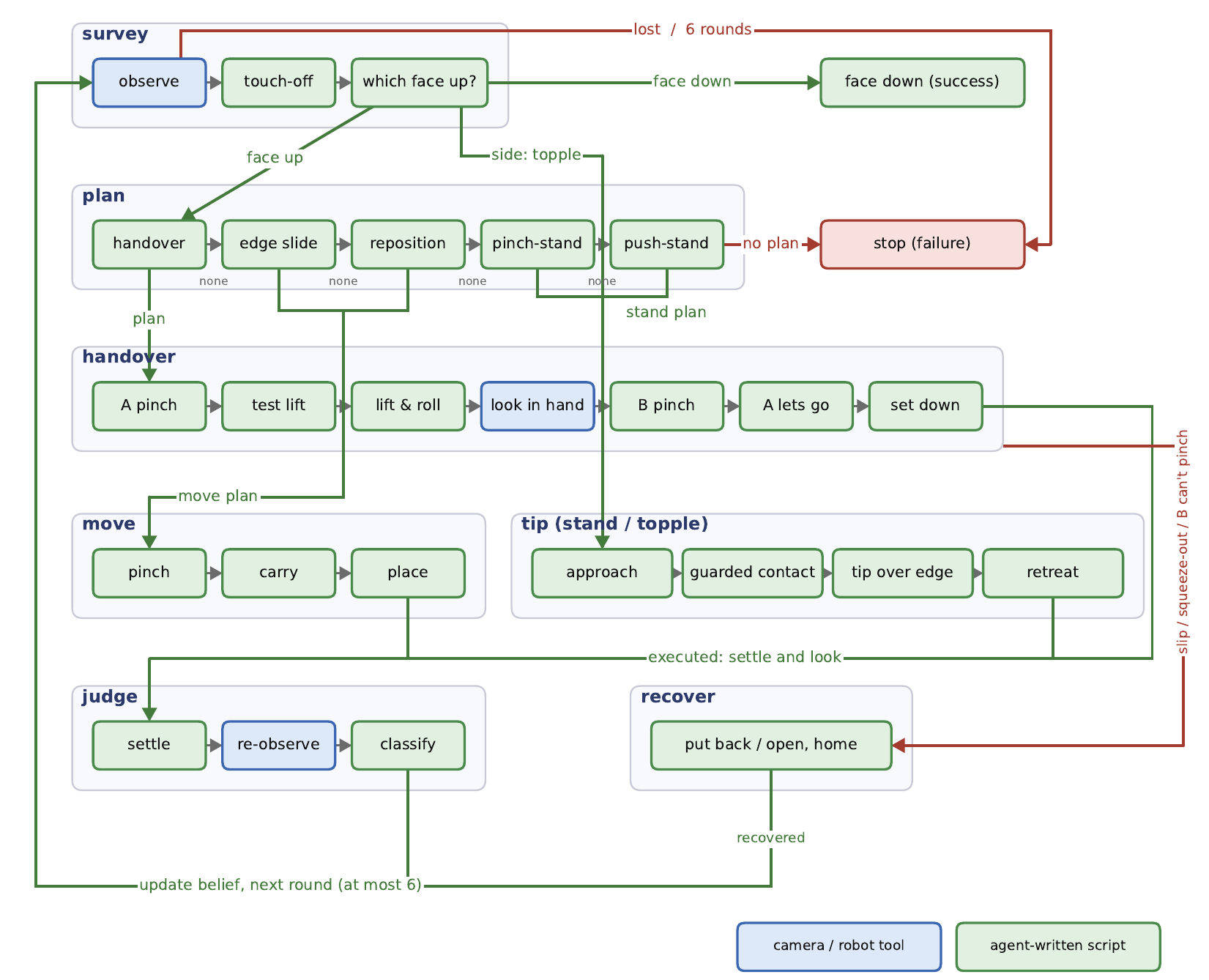}
    \caption{\textbf{Graph policy.} The generated Graph-as-Policy graph routes each observation through plan selection to the strategy components, then re-observes and updates its belief. Blue nodes call the camera or robot; green nodes are agent-written scripts.}
    \label{fig:graph}
\end{figure}

\paragraph{Code Representation Can Generalize Across Manipulation Strategies.}
Qualitative results show that the agent progressively develops distinct manipulation strategies to accommodate different object geometries and constraints, instead of a single grasp-and-rotate motion. For long-handled objects such as the claw hammer, the graph-based code performs an in-air bimanual handover, achieving 14/15 successes. For compact objects such as lock nuts, where a second grasp is difficult, the graph-based code instead uses table-supported manipulation, first standing the object on its edge and then pushing it beyond the tipping point, achieving 15/15 successes on both lock nuts. Wide objects such as multimeters are flipped through coordinated fingertip pushing without requiring a stable grasp, achieving 15/15 successes. When workspace constraints prevent a feasible handover, the graph-based code first repositions the object before attempting the flip, as demonstrated by tape measures (14/15) and pipe wrenches (13/15). Thin objects such as surgical scissors require guarded-contact grasping and slower rotation to prevent slippage, improving from 0/15 successes in earlier graph-based code versions to 13/15 in the final controller. Beyond selecting these strategies, the final code incorporates closed-loop recovery behaviors, including in-hand pose correction, grip verification, and additional pushing when an object only partially flips over.

\begin{figure}[t]
    \centering
    \includegraphics[width=0.72\linewidth]{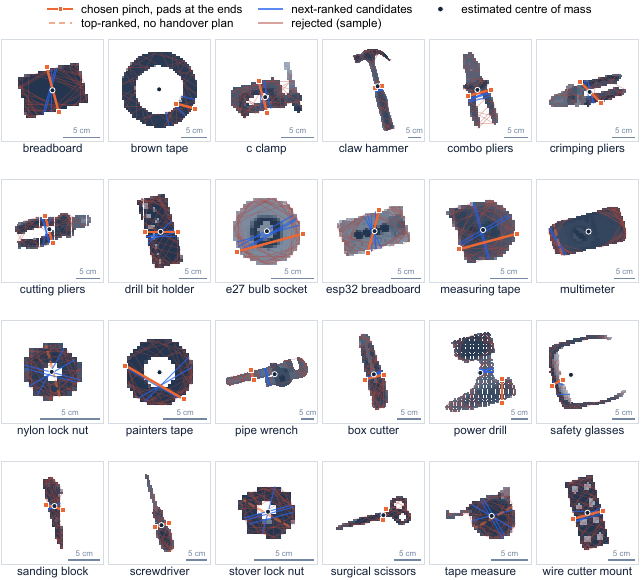}
    \caption{\textbf{Grasp selection for all 24 objects.} Offline run of the policy on the saved episode-0 camera frames. Top-down view of the perceived height map, darker cells higher. Everything shown is computed from the cameras; the object mesh is never read.}
    \label{fig:grasps}
\end{figure}

\begin{figure}[!ht]
    \centering
    \includegraphics[width=\linewidth]{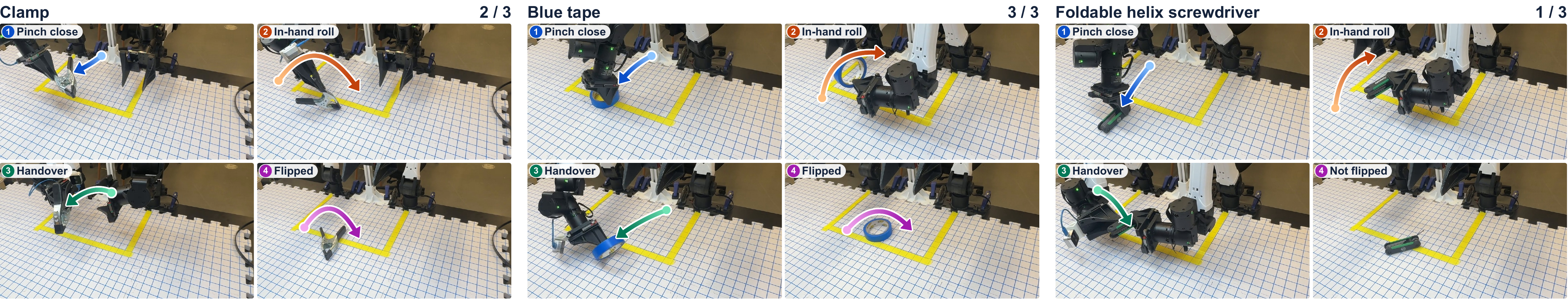}
    \caption{\textbf{Real-world trials.} Keyframes of one real trial per object, with successes out of three trials. The screwdriver trial shown failed. Curves are schematic.}
    \label{fig:real}
\end{figure}

\textbf{Future Work.} We evaluate the FlipAnything code in real-world experiments on three objects whose geometries were not encountered during simulation-based self-learning: a clamp, blue tape, and a foldable helix screwdriver. Without object-specific tuning, the code succeeds in 2 of 3 trials for the clamp, 3 of 3 trials for the blue tape, and 1 of 3 trials for the foldable helix screwdriver. We will study generalization to unseen objects in future work.


\FloatBarrier
\bibliography{refs}

\clearpage

\end{document}

%% file: macros.tex
\newif\ifshowtodos
\showtodostrue

\newif\ifshowadded
\showaddedtrue
\definecolor{addedcolor}{RGB}{0,72,186}

\DeclareCaptionFont{addedfont}{\ifshowadded\color{addedcolor}\fi}

\newlength{\taskpw}

\tikzset{tilechip/.style={overlay, fill=white, fill opacity=0.85, text opacity=1, rounded corners=1pt,
  inner sep=1.1pt, font=\fontsize{5.5}{6}\selectfont}}
\newcommand{\cornertile}[4]{\makebox[#1][c]{%
  \begin{tikzpicture}[inner sep=0pt, baseline=(p.south)]
    \node[anchor=south west] (p) at (0,0) {\includegraphics[width=#1]{#2}};
    \draw[black!25, line width=0.3pt] (p.south west) rectangle (p.north east);
    \if\relax\detokenize{#3}\relax\else
      \node[tilechip, anchor=north west] at ([shift={(1.2pt,-1.2pt)}]p.north west) {#3};\fi
    \if\relax\detokenize{#4}\relax\else
      \node[tilechip, anchor=south west] at ([shift={(1.2pt,1.2pt)}]p.south west) {#4};\fi
  \end{tikzpicture}}}

\definecolor{mjviolet}{RGB}{91,63,166}
\definecolor{sdteal}{RGB}{22,160,132}
\definecolor{vbdamber}{RGB}{214,128,8}

\newlength{\mppw}

\lstdefinestyle{rss}{
  language=Python, basicstyle=\ttfamily\scriptsize, columns=fullflexible, keepspaces=true,
  keywordstyle=\color{blue!55!black}, commentstyle=\color{black!50}\itshape,
  stringstyle=\color{green!35!black}, showstringspaces=false,
  frame=single, rulecolor=\color{black!25}, framesep=3pt, xleftmargin=4pt, xrightmargin=4pt,
  aboveskip=2pt, belowskip=2pt}
\lstdefinestyle{rssout}{
  basicstyle=\ttfamily\scriptsize\color{black!65}, columns=fullflexible, keepspaces=true,
  frame=none, xleftmargin=4pt, aboveskip=1pt, belowskip=1pt}

%% file: refs.bib
@inproceedings{todorov2012mujoco,
  title     = {{MuJoCo}: A physics engine for model-based control},
  author    = {Todorov, Emanuel and Erez, Tom and Tassa, Yuval},
  booktitle = {2012 IEEE/RSJ International Conference on Intelligent Robots and Systems (IROS)},
  pages     = {5026--5033},
  year      = {2012},
  month     = oct,
  publisher = {IEEE},
  doi       = {10.1109/IROS.2012.6386109},
}

@inproceedings{chavandafle2014extrinsic,
  author = {Chavan-Dafle, N. and Rodriguez, A. and Paolini, R. and Tang, B. and Srinivasa, S. S. and Erdmann, M. and Mason, M. T. and Lundberg, I. and Staab, H. and Fuhlbrigge, T.},
  title = {{Extrinsic Dexterity: In-Hand Manipulation with External Forces}},
  booktitle = {IEEE International Conference on Robotics and Automation (ICRA)},
  year = {2014},
  url = {https://personalrobotics.cs.washington.edu/publications/chavandafle2014extrinsic/},
}

@inproceedings{liang2023codeaspolicies,
  author = {Liang, Jacky and Huang, Wenlong and Xia, Fei and Xu, Peng and Hausman, Karol and Ichter, Brian and Florence, Pete and Zeng, Andy},
  title = {{Code as Policies: Language Model Programs for Embodied Control}},
  booktitle = {IEEE International Conference on Robotics and Automation (ICRA)},
  year = {2023},
  url = {https://arxiv.org/abs/2209.07753},
}

@misc{chen2026gap,
  author = {Chen, Kaiyuan and Xie, Shuangyu and Fu, Letian and Yu, Justin and Pacini, William and Bajamahal, Sandeep and Kim, Hudson and Drake, Jaimyn and Kim, Daehwa and Xue, Haoru and Francis, Jonathan and Juette, Christian and Schaldenbrand, Peter and Seker, Muhammet Yunus and Wickramarachchi, Ruwan and Yoo, Uksang and Wang, Guanzhi and Murali, Adithyavairavan and Sundaralingam, Balakumar and Sastry, S. Shankar and Zhu, Yuke and Fan, Linxi Jim and Huang, Spencer and Goldberg, Ken},
  title = {{GaP: A Graph-as-Policy Multi-Agent Self-Learning Harness For Variational Automation Tasks}},
  year = {2026},
  eprint = {2607.05369},
  archivePrefix = {arXiv},
  primaryClass = {cs.RO},
  doi = {10.48550/arXiv.2607.05369},
  url = {https://arxiv.org/abs/2607.05369},
}

@misc{anthropic2026fable,
  author = {{Anthropic}},
  title = {{Introducing Claude Fable 5.1 and Claude Mythos 5.1}},
  year = {2026},
  month = {September},
  url = {https://www.anthropic.com/claude-fable-and-mythos-5-1},
  note = {Accessed October 26, 2026},
}

@misc{chen2026robosimstudio,
  author = {Chen, Kaiyuan and Xie, Shuangyu and Yoo, Uksang and Zhu, Xinyue and Magner, Cael and Zhu, Alvin and Kim, Daehwa and Heiden, Eric and Coumans, Erwin and Macklin, Miles and Huang, Spencer and Goldberg, Ken},
  title = {{RoboSimStudio: An Interactive Toolkit for Human and Agent Rapid Real2Sim of Robot Manipulation Tasks}},
  year = {2026},
  eprint = {2610.18835},
  archivePrefix = {arXiv},
  primaryClass = {cs.RO},
  doi = {10.48550/arXiv.2610.18835},
  url = {https://arxiv.org/abs/2610.18835},
}

@article{wei2022coacd,
  author = {Wei, Xinyue and Liu, Minghua and Ling, Zhan and Su, Hao},
  title = {{Approximate Convex Decomposition for 3D Meshes with Collision-Aware Concavity and Tree Search}},
  journal = {ACM Transactions on Graphics},
  volume = {41},
  number = {4},
  articleno = {42},
  pages = {1--18},
  year = {2022},
  doi = {10.1145/3528223.3530103},
  url = {https://doi.org/10.1145/3528223.3530103},
}

@inproceedings{hou2018pivoting,
  author = {Hou, Yifan and Jia, Zhenzhong and Mason, Matthew T.},
  title = {{Fast Planning for 3D Any-Pose-Reorienting using Pivoting}},
  booktitle = {2018 IEEE International Conference on Robotics and Automation (ICRA)},
  pages = {1631--1638},
  year = {2018},
  url = {https://www.ri.cmu.edu/app/uploads/2018/03/ICRA2018_Reorienting_by_Pivoting.pdf},
}

@inproceedings{zhou2023ungraspable,
  author = {Zhou, Wenxuan and Held, David},
  title = {{Learning to Grasp the Ungraspable with Emergent Extrinsic Dexterity}},
  booktitle = {Proceedings of the 6th Conference on Robot Learning},
  series = {Proceedings of Machine Learning Research},
  volume = {205},
  pages = {150--160},
  year = {2023},
  publisher = {PMLR},
  url = {https://proceedings.mlr.press/v205/zhou23a.html},
}

@inproceedings{yang2024extrinsic,
  author = {Yang, Shih-Min and Magnusson, Martin and Stork, Johannes A. and Stoyanov, Todor},
  title = {{Learning Extrinsic Dexterity with Parameterized Manipulation Primitives}},
  booktitle = {2024 IEEE International Conference on Robotics and Automation (ICRA)},
  pages = {5404--5410},
  year = {2024},
  organization = {IEEE},
  url = {https://arxiv.org/abs/2310.17785},
}

@article{fu2026capx,
  title={CaP-X: A Framework for Benchmarking and Improving Coding Agents for Robot Manipulation},
  author={Fu, Max and Yu, Justin and El-Refai, Karim and Kou, Ethan and Xue, Haoru and Huang, Huang and Xiao, Wenli and Wang, Guanzhi and Li, Fei-Fei and Shi, Guanya and others},
  journal={arXiv preprint arXiv:2603.22435},
  year={2026}
}

@inproceedings{yamada2025combograsp,
  author = {Yamada, Jun and Mitchell, Alexander Luis and Collins, Jack and Posner, Ingmar},
  title = {{COMBO-Grasp: Learning Constraint-Based Manipulation for Bimanual Occluded Grasping}},
  booktitle = {Proceedings of the 9th Conference on Robot Learning},
  series = {Proceedings of Machine Learning Research},
  volume = {305},
  pages = {1102--1119},
  year = {2025},
  publisher = {PMLR},
  url = {https://proceedings.mlr.press/v305/yamada25a.html},
}
